\documentclass[12pt]{article}
\usepackage{arxiv}
\usepackage{graphicx} % Required for inserting images
\usepackage{tikz}
\usepackage{url}
\usepackage{booktabs}
\usepackage{subcaption}
\usepackage{adjustbox}
\usepackage{multirow,multicol}
\usepackage{amsmath,amsfonts,amssymb}
\usepackage{algorithm}
\usepackage{algorithmic}
\usepackage{amsmath,amsfonts,bm}

\def\eqref#1{equation~\ref{#1}}
\def\1{\bm{1}}

\def\vdelta{{\bm{\delta}}}
\def\va{{\bm{a}}}

\def\ve{{\bm{e}}}

\def\vg{{\bm{g}}}

\def\vm{{\bm{m}}}

\def\vu{{\bm{u}}}
\def\vv{{\bm{v}}}

\def\vx{{\bm{x}}}

\DeclareMathAlphabet{\mathsfit}{\encodingdefault}{\sfdefault}{m}{sl}
\SetMathAlphabet{\mathsfit}{bold}{\encodingdefault}{\sfdefault}{bx}{n}

\DeclareMathOperator*{\argmax}{arg\,max}
\DeclareMathOperator*{\argmin}{arg\,min}

\title{Empirical evaluation of the Frank-Wolfe methods for constructing white-box adversarial attacks}
\newcommand{\shorttitle}{Evaluation of the Frank-Wolfe methods for white-box attacks}

\author{Kristina Korotkova \\
Moscow Institute of Physics and Technology\\
       Moscow, Russia \\
       \url{korotkova.km@phystech.edu} 
       \And
       Aleksandr Katrutsa\\
       Skoltech \\
       Moscow, Russia \\
       \url{amkatrutsa@gmail.com}
       } 

\begin{document}

\maketitle

\begin{abstract}
The construction of adversarial attacks for neural networks appears to be a crucial challenge for their deployment in various services.
To estimate the adversarial robustness of a neural network, a fast and efficient approach is needed to construct adversarial attacks.
Since the formalization of adversarial attack construction involves solving a specific optimization problem, we consider the problem of constructing an efficient and effective adversarial attack from a numerical optimization perspective.
Specifically, we suggest utilizing advanced projection-free methods, known as modified Frank-Wolfe methods, to construct white-box adversarial attacks on the given input data.
We perform a theoretical and numerical evaluation of these methods and compare them with standard approaches based on projection operations or geometrical intuition.
Numerical experiments are performed on the MNIST and CIFAR-10 datasets, utilizing a multiclass logistic regression model, the convolutional neural networks (CNNs), and the Vision Transformer (ViT).
\end{abstract}

\section{Introduction}
\label{sec:intro}

The vulnerability of deep neural networks (DNNs) to adversarial attacks has been widely recognized for over a decade~\cite{szegedy2013intriguing,goodfellow2014explaining} and remains a critical challenge for the security of modern machine learning models. 
As artificial intelligence technologies advance, ensuring the robustness and reliability of such systems becomes increasingly important. 
Adversarial attacks, which intentionally introduce small and often imperceptible perturbations to input data, pose a significant threat by inducing erroneous model predictions. Consequently, it is essential not only to develop new defenses but also to analyze and understand the vulnerabilities of existing models.
In this study, we focus on the image classification task, where perturbation is explicitly formalized as pixel-level noise, making performance comparisons straightforward. 
We consider white-box attacks, i.e., those that use both model outputs and gradients.

A standard approach to formalizing adversarial example generation is to maximize the task loss with respect to a perturbation subject to a norm-ball constraint. 
A fair analysis of such formulations across norms and a comparison of corresponding optimization methods are essential for building efficient attacks. 
This work empirically compares projection-free methods with popular baselines under different norm constraints that control perturbation magnitude. 
Since different norms induce perturbations with different structures, identifying suitable optimization methods per norm accelerates robustness evaluation.

Although the standard choice of constraint is the $\ell_\infty$ norm, which typically yields dense, almost uniformly nonzero perturbations, it is well-known~\cite{candes2014mathematics} that $\ell_1$-constrained problems tend to have sparse solutions. 
Thus, the $\ell_1$-ball constraint can lead to sparse adversarial perturbations. 
In this case, projection-based methods such as FGSM~\cite{goodfellow2014explaining} or PGD~\cite{madry2017towards} may be slower due to the more computationally expensive projection step. 
This motivates projection-free (Frank--Wolfe-type) methods, which avoid projections and can converge faster; moreover, advanced Frank--Wolfe variants can further improve attack quality and efficiency.

We evaluate the optimization methods to generate adversarial attacks on pre-trained models. 
We start with logistic regression, proceed to a convolutional neural network, and conclude with a Vision Transformer (ViT)~\cite{dosovitskiy2020image}. 
Our results highlight the potential of projection-free methods and provide practical recommendations on when to prefer them.

The main contributions are:
\begin{enumerate}
  \item We study several projection-free methods for generating adversarial attacks under $\ell_1$, $\ell_2$, and $\ell_\infty$ constraints.
  \item We analyze the empirical performance and properties of the resulting perturbations, identifying effective choices for each norm/model class.
  \item  We analyze the sparsity of the adversarial attacks and discuss the related features of the considered methods.
\end{enumerate}

\paragraph{Related work.}
Adversarial attacks threaten a wide range of applications beyond images. 
Recommender systems can be manipulated~\cite{deldjoo2021survey,cao2020adversarial,di2020taamr}; conversational agents can be misled~\cite{shang2024adversarial}; speech systems can be compromised~\cite{zelasko2021adversarial,qin2019imperceptible,korzh2025certification}; and machine translation can be attacked~\cite{chertkov2024translate}. 
Financial decision systems, including credit scoring~\cite{ghamizi2020search,kumar2021evolutionary} and fraud detection~\cite{lunghi2024assessing,gupta2025adversarial}, are also vulnerable.

Early adversarial methods relied on projected gradient descent and its variants~\cite{madry2017towards,goodfellow2014explaining,kurakin2016adversarial}. 
Later methods exploit decision boundary geometry~\cite{moosavi2016deepfool}, tailored loss functions~\cite{carlini2017towards}, or domain-specific insights~\cite{narodytska2016simple}. 
Universal perturbations~\cite{moosavi2017universal} applicable to any sample from the given dataset have been extensively studied in~\cite{khrulkov2018art,kuvshinova2025sparse,zhang2021survey,chen2020universal}. 
Comprehensive surveys include~\cite{zhou2022adversarial,costa2024deep,zhang2024adversarial}.

The Frank--Wolfe method~\cite{frank1956algorithm,pokutta2024frank} and its modifications~\cite{bomze2024frank} are well-known in machine learning~\cite{jaggi2013revisiting,lacoste2015global,katrutsa2015stress}, but their use in adversarial attacks is limited. Study~\cite{chen2020frank} considers the vanilla Frank--Wolfe attack, omitting advanced variants. 
Our study fills this gap with a systematic evaluation of advanced Frank--Wolfe modifications for adversarial example generation.

\section{Problem statement}
\label{sec:problem}

Let $f_{\boldsymbol{\theta}}$ be a pre-trained neural network and $\vx\in\mathbb{R}^d$ be an input. 
We consider multiclass image classification, so $d$ is the number of pixels (after vectorization). 
Assume $f_{\boldsymbol{\theta}}$ correctly predicts the true label $y_{\mathrm{true}}$ for $\vx$. 
An adversarial example for $\vx$ is $\vx+\vdelta$ such that the predicted label $y_\delta \neq y_{\mathrm{true}}$, where $\vdelta \in\mathbb{R}^d$ is a small perturbation. 
Visual similarity is enforced by $\|\vdelta\|_p \le \varepsilon$ for a given $\varepsilon>0$, typically aligned with pixel ranges (see Section~\ref{sec:exp}).

A common approach to search proper perturbation $\vdelta^*$ is to maximize the training loss $\mathcal{L}$ under a norm-ball constraint:
\begin{equation}
\begin{split}
& \vdelta^* \in \argmax_{\vdelta} \ \mathcal{L}(\vx+\vdelta\mid f_{\boldsymbol{\theta}}),\\
\text{subject to}\quad & \|\vdelta\|_p \le \varepsilon.
\end{split}
\label{eq:basic_problem}
\end{equation}
We focus on per-instance perturbations (not universal attacks~\cite{zhang2021survey}) and use cross-entropy loss as $\mathcal{L}$.

While $\varepsilon$ is data-dependent, the choice of norm should be paired with a suitable optimizer for solving~(\ref{eq:basic_problem}). 
Standard norms are $\ell_2$, $\ell_1$, and $\ell_\infty$, each constraining magnitude differently. 
White-box baselines typically rely on projected gradient methods (e.g., FGSM~\cite{goodfellow2014explaining}, PGD~\cite{madry2017towards}), which require computing the projection at each iteration:
\begin{equation}
    \vdelta_{\mathrm{proj}} \in \argmin_{\vdelta\in \mathcal{S}} \ \|\vu_k - \vdelta\|_2,
    \label{eq:proj_general}
\end{equation}
where $\mathcal{S}=\{ \vdelta \mid \|\vdelta\|_p\le \varepsilon\}$. 
Closed-form solutions for problem~(\ref{eq:proj_general}) exist for $p\in\{2,\infty\}$, but not for $p=1$, which requires specialized procedures~\cite{duchi2008efficient,condat2016fast}. 
In addition, $\ell_1$ constraints often yield sparse $\vdelta^*$, revealing sensitive pixels.

Projection-free (Frank--Wolfe) methods~\cite{frank1956algorithm,jaggi2013revisiting,pokutta2024frank} avoid projections by solving the following auxiliary optimization problem in every iteration
\begin{equation}
    \vv^* \in \argmin_{\vv\in\mathcal{S}} \ \big\langle \nabla_{\vdelta}\mathcal{L}(\vx+\vdelta_k \mid f_{\boldsymbol{\theta}}), \ \vv\big\rangle,
    \label{eq:lmo}
\end{equation}
and updating
\begin{equation}
    \vdelta_{k+1} = (1-\gamma_k)\,\vdelta_k + \gamma_k\,\vv^*,\qquad \gamma_k\in(0,1).
    \label{eq:fw-update}
\end{equation}
The subproblem~(\ref{eq:lmo}) is called the linear minimization oracle (LMO)~\cite{lan2013complexity,thuerck2023learning,juditsky2016solving}. 
For $\mathcal{S}$ induced by $p\in\{1,2,\infty\}$, problem (\ref{eq:lmo}) has closed-form solutions, making projection-free methods attractive for white-box attacks. 
However, most prior works~\cite{chen2020frank,tsiligkaridis2022understanding} use only the vanilla Frank-Wolfe method. 
We evaluate several advanced variants across models and norms to fill this gap.

In the next section, we outline the Frank--Wolfe method and its most promising modifications, discussing theory and practice for~(\ref{eq:basic_problem}), and show that advanced variants substantially improve performance, especially under $\ell_1$ constraints.

\section{Frank--Wolfe method modifications}
\label{sec:fwmods}

The Frank-Wolfe method addresses constrained problems whose feasible sets admit analytical solution of the LMO subproblem~(\ref{eq:lmo}). 
Starting from $\vdelta_0$, FW repeats~(\ref{eq:lmo}) and~(\ref{eq:fw-update}). 
A standard choice for stepsize is $\gamma_k = 2/(k+2)$, and stopping criteria include objective stabilization or the duality gap. 
% For convex objectives, the convergence rate is $O(1/T)$ after $T$ iterations; for nonconvex objectives, $O(1/\sqrt{T})$.
% In our setting, convexity holds only for the logistic regression model. 
% Hence, empirical results are crucial. 
For the reader's convenience, we summarize in Table~\ref{tab:lmo_solutions} the LMO solutions under the considered norms.

\begin{table}[htbp]
\centering
  \caption{Closed-form LMO solutions for $\ell_1$, $\ell_2$, and $\ell_\infty$ balls with radius $\varepsilon$. Here $\vg$ denotes the current gradient of the objective function.}
  {
    \begin{tabular}{ll}
      \toprule
      \textbf{Feasible set $\mathcal{S}$} & \textbf{LMO solution} \\
      \midrule
      $\{\vv:\|\vv\|_\infty \le \varepsilon\}$ &
      $\vv^* \,=\, -\,\varepsilon\,\mathrm{sign}(\vg)$ \\
      $\{\vv:\|\vv\|_2 \le \varepsilon\}$ &
      $\vv^* \,=\, -\,\varepsilon\, \dfrac{\vg}{\|\vg\|_2}$ \\
      $\{\vv:\|\vv\|_1 \le \varepsilon\}$ &
      $\vv^* \,=\, -\,\varepsilon\, \mathrm{sign}(g_{i^*})\,\ve_{i^*}, \quad i^* \in \arg\max_i |g_i|$ \\
      \bottomrule
    \end{tabular}
    \label{tab:lmo_solutions}
  }
\end{table}

However, multiple modifications aim to accelerate the convergence of the vanilla FW method~\cite{bomze2024frank}.
In the following sections, we briefly introduce the most promising ones and discuss their key properties.

\subsection{Momentum Frank--Wolfe}
The momentum Frank-Wolfe (FWm) method augments the FW step with a momentum term to reduce zig-zagging behaviour and accelerate convergence, especially in ill-conditioned problems~\cite{chen2022multistep}. 
Let $\vm_k$ be the momentum accumulator:
\[
\vm_{k+1} = \beta \vm_k + (1-\beta) \vg_k, \qquad \beta\in(0,1),
\]
then every iteration of FWm makes the following update
\[
\vv_k \in \argmin_{\vv\in\mathcal{S}} \langle \vv, \vm_{k+1}\rangle,\qquad
\vdelta_{k+1} = \vdelta_k + \gamma_k(\vv_k - \vdelta_k).
\]
Under standard smoothness assumptions, FWm achieves $O(1/\sqrt{T})$ sublinear convergence in the nonconvex setting~\cite{chen2020frank}, matching the convergence rate of classical FW, but often appears faster in practice.

\subsection{Away-steps Frank-Wolfe and pairwise Frank-Wolfe methods}
The other two modifications of the vanilla Frank-Wolfe method are the away-steps Frank-Wolfe (AFW) and the pairwise Frank-Wolfe (PFW) methods.
To describe them uniformly, one can represent the current solution approximation~$\vx_k$ as a convex combination of \emph{atoms} $\va_i \in \mathcal{A}_k$, where $\mathcal{A}_k \subset \mathcal{S}$ denotes an active set in the $k$-th iteration.
In particular, FW-type methods maintain an active set $\mathcal{A}_k$ of atoms and the corresponding coefficients $\{\alpha_{i}^{(k)}\}$ such that $\sum_{i} \alpha_{i}^{(k)}=1$, $\alpha_{i}^{(k)}\ge 0$, and
\[
\vx_k = \sum_{i=1} ^{|\mathcal{A}_k|}\alpha_{i}^{(k)} \va_i.
\]

AFW accelerates FW for convex problems by allowing ``away steps'' that reduce the weight of suboptimal atoms in the current active set. 
If the FW step offers limited improvement, AFW can step away from the worst active atom, enabling faster pruning of poor atoms and improved conditioning.

PFW performs pairwise mass transfers between the worst active atoms and the best FW atoms, directly redistributing weight to refine the convex combination more efficiently. 
This often yields sparser and more compact representations~\cite{lacoste2015global}.

For the reader's convenience, we summarize in Table~\ref{tab:active_set} the interpretation of a single FW-type iteration for the FW, AFW, and PFW methods, from the perspective of updates to the active set and the corresponding weights.

% \paragraph{Theory.} Under strong convexity and smoothness, and when the feasible set is a polytope, both AFW and PFW enjoy global linear convergence on “good” steps \cite{lacoste2015global}:
% \[
% L(\vx^T) - L^* \ \le\ (1-\rho)^k \,\big[L(\vx^0)-L^*\big],
% \]
% with $\rho = \frac{\mu}{4L}\left(\frac{\delta}{M}\right)^2$ depending on strong convexity $\mu$, smoothness $L$, the pyramidal width $\delta$, and the polytope diameter $M$.

% Table~\ref{tab:active_set} sketches the update patterns for FW, AFW, and PFW.

\begin{table}[htbp]
\centering
  \caption{The update of active set and atoms' coefficients implemented in FW, AFW, and PFW methods.}
  {%
    \begin{tabular}{c p{8.5cm}}
      \toprule
      \textbf{Algorithm} & \textbf{Active set and coefficient update (schematic)} \\
      \midrule
      FW &
      $
      \begin{aligned}
        &\text{If } s_t \notin \mathcal{S}_t:\ \mathcal{S}_{t+1}\!\leftarrow\!\mathcal{S}_t \cup \{s_t\},\ \alpha_{s_t}\!\leftarrow\!0 \\
        &\forall v\in \mathcal{S}_{t+1}:\ \alpha_v^{(t+1)} = (1-\gamma)\,\alpha_v^{(t)} \\
        &\alpha_{s_t}^{(t+1)} \mathrel{+}= \gamma \\
        &\text{If } \gamma=1:\ \mathcal{S}_{t+1}=\{s_t\},\ \alpha_{s_t}^{(t+1)}=1
      \end{aligned}
      $ \\
      \midrule
      AFW &
      $
      \begin{aligned}
        &\text{If FW step: same as FW} \\
        &\text{If Away step from } v_t\in \mathcal{S}_t: \\
        &\qquad \forall v\in \mathcal{S}_t:\ \alpha_v^{(t+1)} = (1+\gamma)\,\alpha_v^{(t)} \\
        &\qquad \alpha_{v_t}^{(t+1)} \mathrel{-}= \gamma \\
        &\qquad \text{If } \alpha_{v_t}^{(t+1)}=0:\ \mathcal{S}_{t+1}\!\leftarrow\!\mathcal{S}_t\setminus\{v_t\}
      \end{aligned}
      $ \\
      \midrule
      PFW &
      $
      \begin{aligned}
        &s_t \in \arg\min_{v}\ \langle \nabla f(x_t), v\rangle,\quad
          v_t \in \arg\max_{v\in \mathcal{S}_t}\ \langle \nabla f(x_t), v\rangle \\
        &\text{(add } s_t \text{ if needed)},\ \alpha_{s_t}\!\leftarrow\!0 \\
        &\alpha_{s_t}^{(t+1)} \mathrel{+}= \gamma,\quad \alpha_{v_t}^{(t+1)} \mathrel{-}= \gamma \\
        &\text{If } \alpha_{v_t}^{(t+1)}=0:\ \mathcal{S}_{t+1}\!\leftarrow\!\mathcal{S}_t\setminus\{v_t\} \\
        &\text{If } \alpha_{s_t}^{(t+1)}=1:\ \mathcal{S}_{t+1}=\{s_t\}
      \end{aligned}
      $ \\
      \bottomrule
    \end{tabular}
    \label{tab:active_set}
  }
\end{table}

\section{Numerical experiments}
\label{sec:exp}

We compare adversarial attack methods across datasets and models, reporting test accuracy after attacks, average runtime per image, and average number of nonzero pixels in the perturbation. 
Visual examples illustrate key phenomena.

\paragraph{Models.}

Architectures include: logistic regression on MNIST, ResNet-56, and ViT on CIFAR-10. 
Logistic regression is trained from scratch; ResNet-56 is pre-trained \cite{chenyaofo_pytorch_cifar_models} and ViT is fine-tuned. 
Key characteristics are presented in Table~\ref{tab:architectures}.

\begin{table}[htbp]
\centering
  \caption{Description of the considered models.}
  {%
    \begin{tabular}{lcccc}
      \toprule
      \textbf{Model} & \textbf{Layers} & \textbf{Params (M)} & \textbf{Size (MB)} & \textbf{Test acc. (\%)} \\
      \midrule
      LogReg & 1 & 0.008 & $<$1 & 92.68 \\
      ResNet56 & 56 & 0.86 & 3.4 & 94.37 \\
      % VGG19\_bn & 19 & 20.57 & 78.5 & 93.91 \\
      ViT & 12 & 85.81 & 327.4 & 97.28 \\
      \bottomrule
    \end{tabular}
    \label{tab:architectures}
  }
\end{table}

\paragraph{Datasets.}
We use the MNIST and CIFAR-10 datasets. 
The first dataset is grayscale, whereas CIFAR-10 is colored and more diverse, enabling comparisons across complexity levels.
To compare the performance of the considered optimizers, we use the test subset of every dataset.

\subsection{Sparse perturbations under $\ell_1$ constraint}

\begin{figure}[htbp]
\centering
  {%
  \begin{subfigure}{0.3\textwidth}
\includegraphics[width=\textwidth]{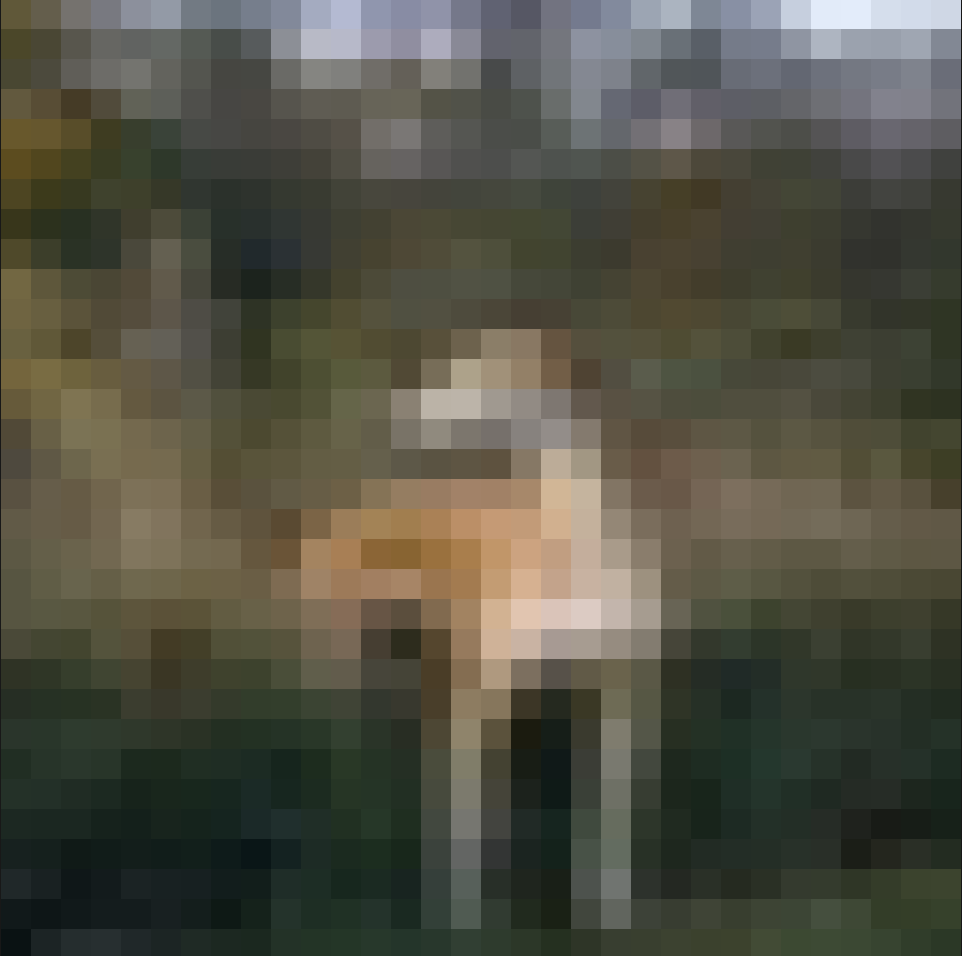}
    \subcaption{Original}
    \end{subfigure}
    ~
    \begin{subfigure}{0.3\textwidth}
{\includegraphics[width=\textwidth]{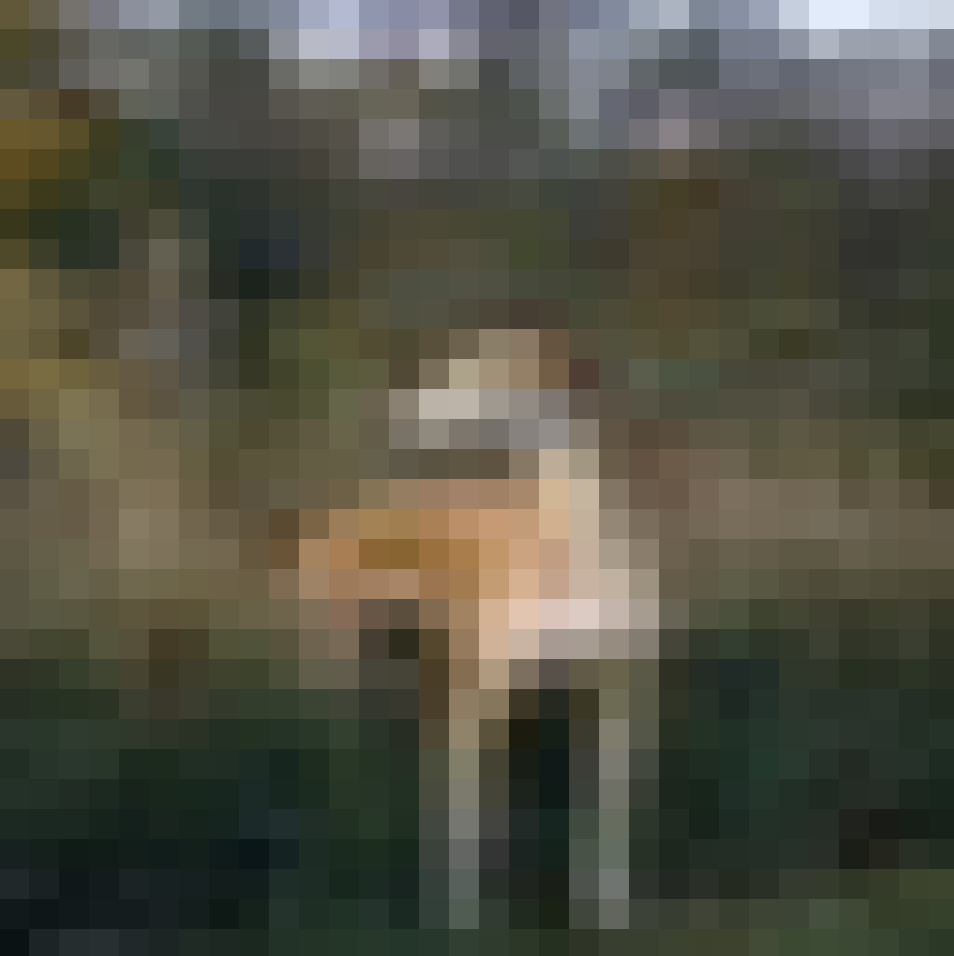}}
    \subcaption{Adversarial}
    \end{subfigure}
    ~
    \begin{subfigure}{0.3\textwidth}
\includegraphics[width=\textwidth]{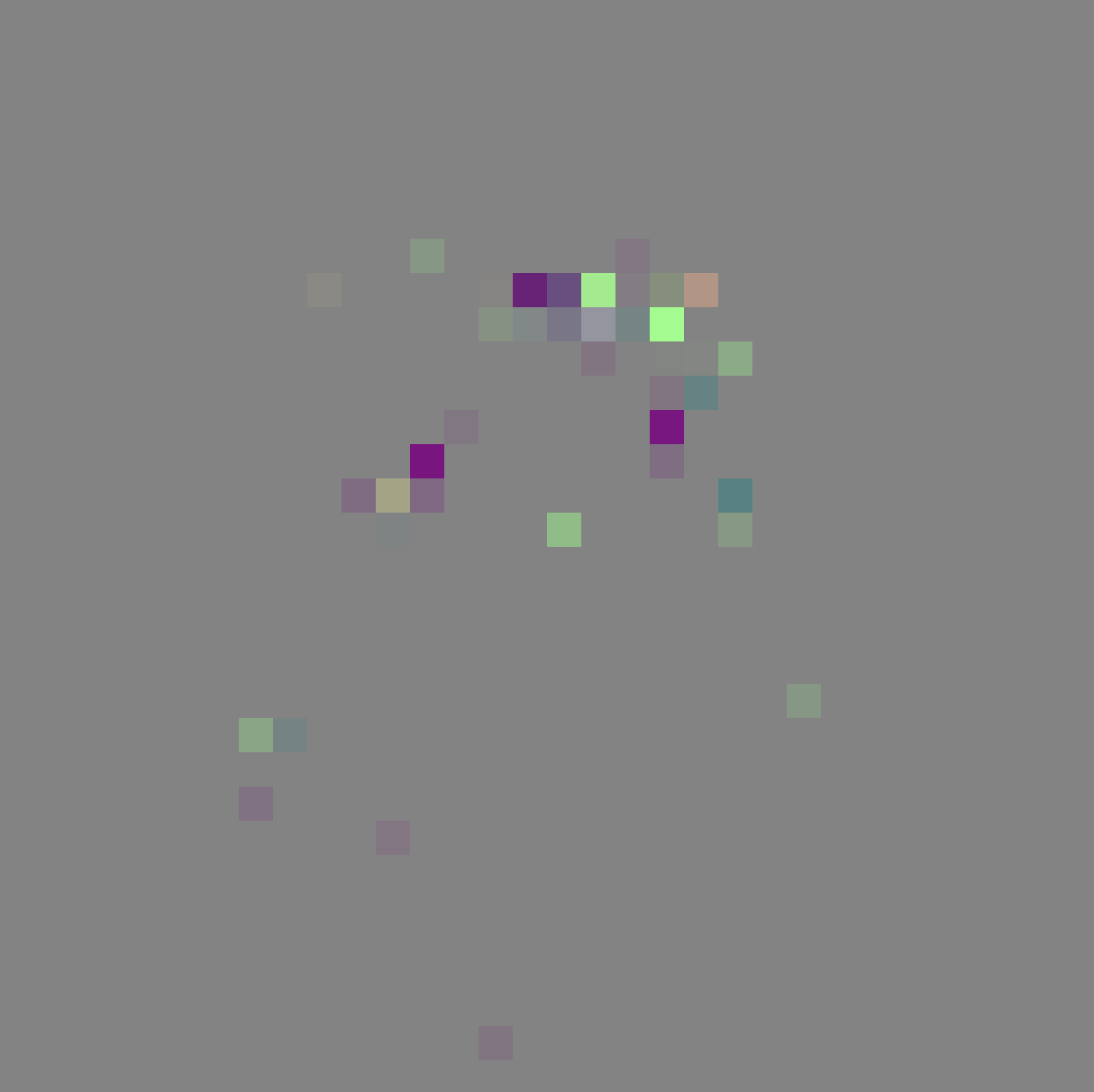}
    \subcaption{Perturbation}
    \end{subfigure}
  }
  \caption{FW attack (15 steps), $\ell_1$-ball, $\varepsilon=64/255$, VGG-19, CIFAR-10. Left: original (“deer”); middle: adversarial (“bird”); right: perturbation.}
   \label{fig:cifar10_fw_l_1}
\end{figure}

\begin{figure}[htbp]
  {%
  \begin{subfigure}{0.3\textwidth}
    \includegraphics[width=\textwidth]{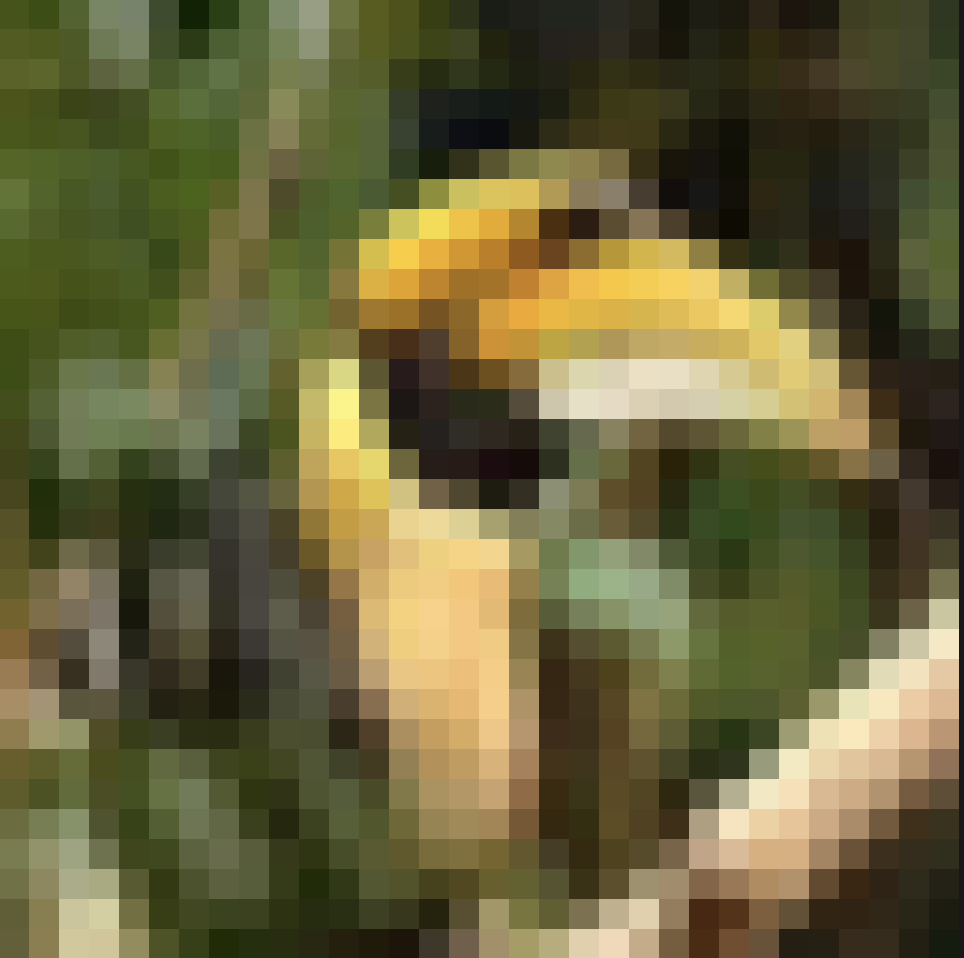}
    \subcaption{Original}
    \end{subfigure}
    ~
    \begin{subfigure}{0.3\textwidth}
    \includegraphics[width=\textwidth]{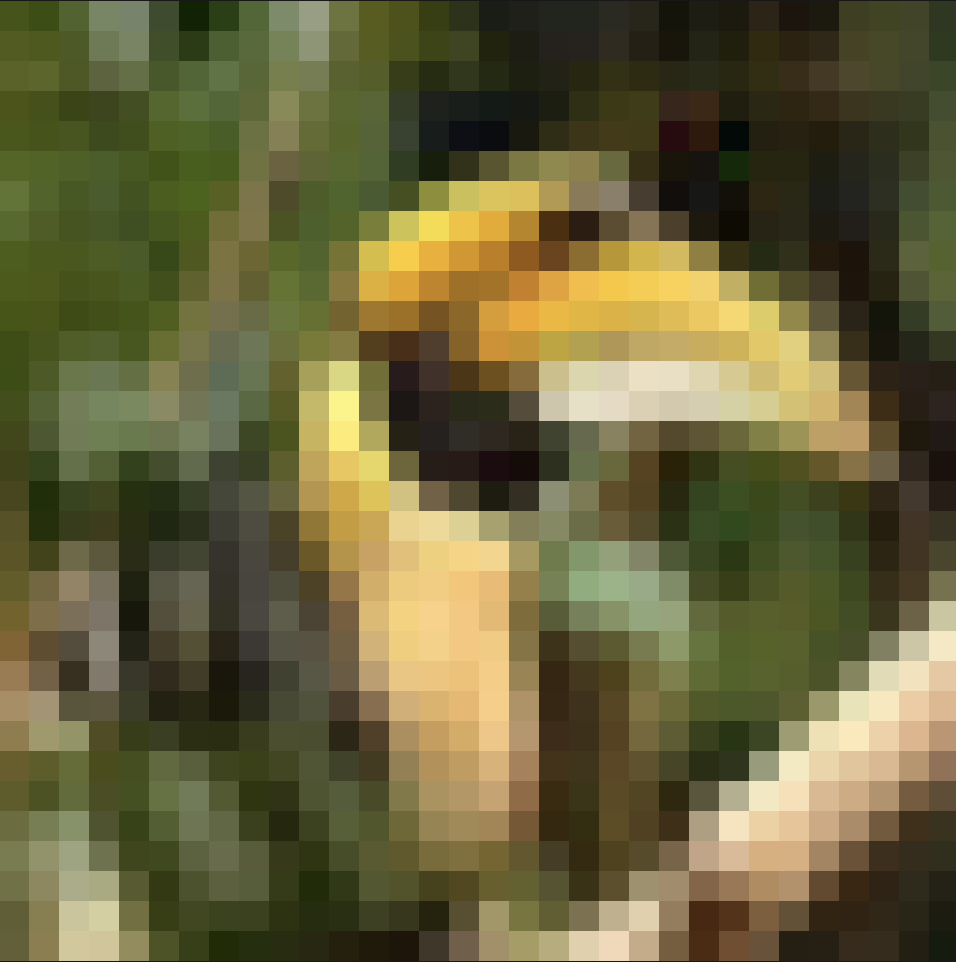}
    \subcaption{Adversarial}
    \end{subfigure}
    ~
    \begin{subfigure}{0.3\textwidth}
    \includegraphics[width=\textwidth]{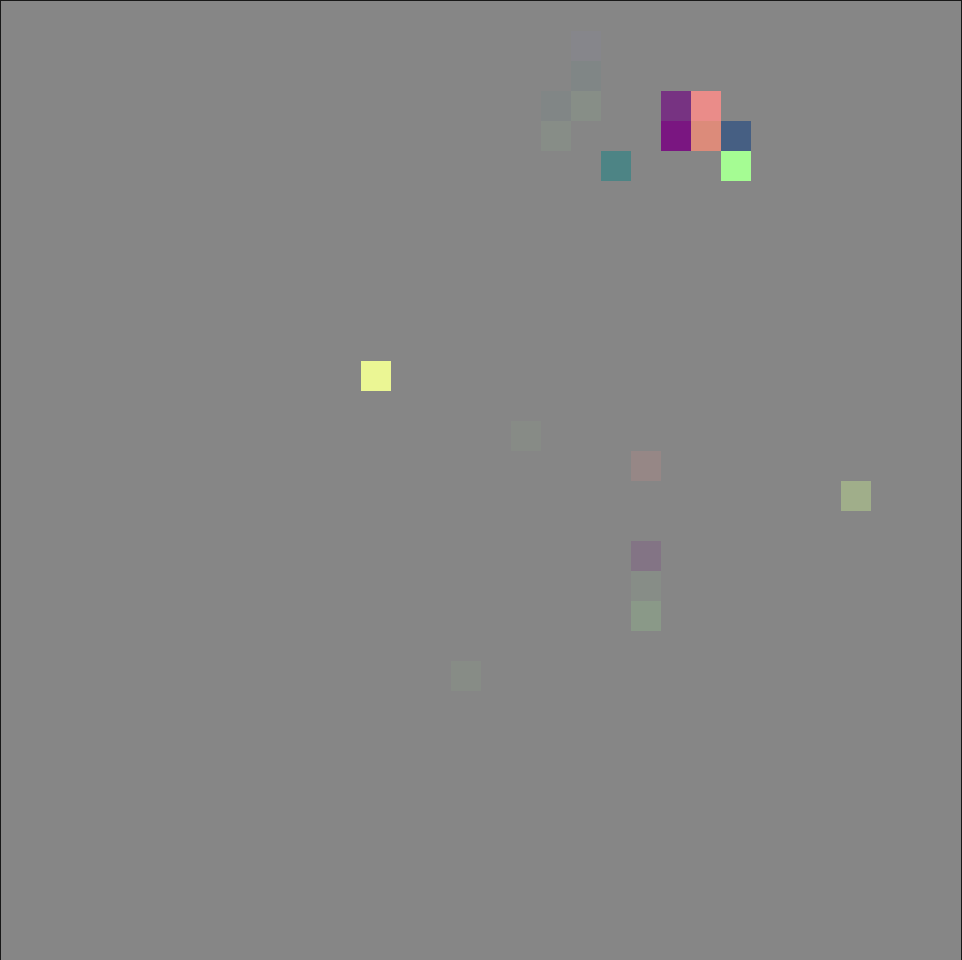}
    \subcaption{Perturbation}
    \end{subfigure}
  }
  \caption{AFW attack (15 steps), $\ell_1$-ball, $\varepsilon=64/255$, VGG-19, CIFAR-10. Left: original (“bird”); middle: adversarial (“frog”); right: perturbation.}
  \label{fig:cifar10_afw_l_1}
\end{figure}

\begin{figure}[htbp]
  {%
  \begin{subfigure}{0.3\textwidth}
      \includegraphics[width=\linewidth]{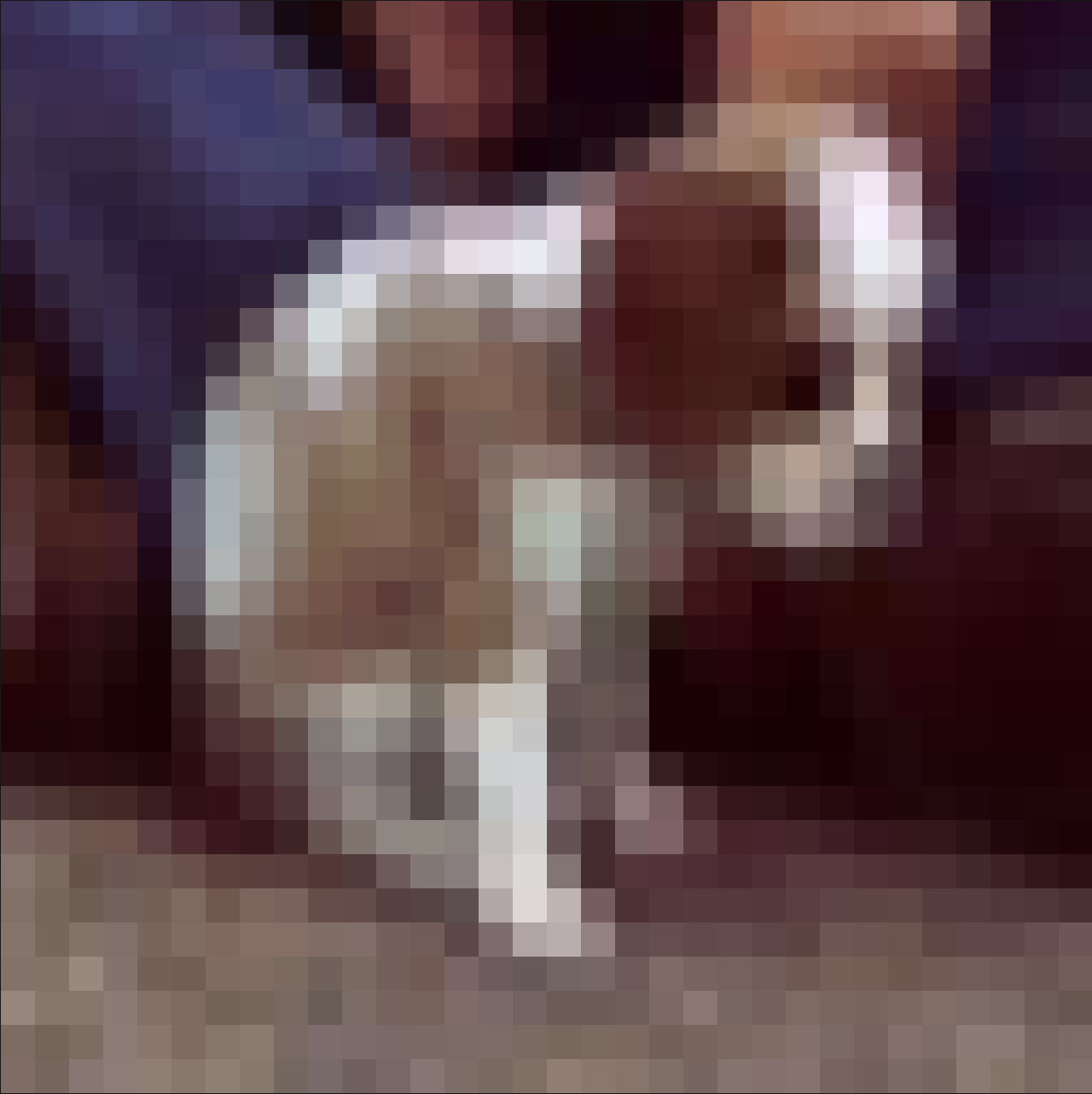}
      \subcaption{Original}
  \end{subfigure}
    ~
    \begin{subfigure}{0.3\textwidth}
        \includegraphics[width=\linewidth]{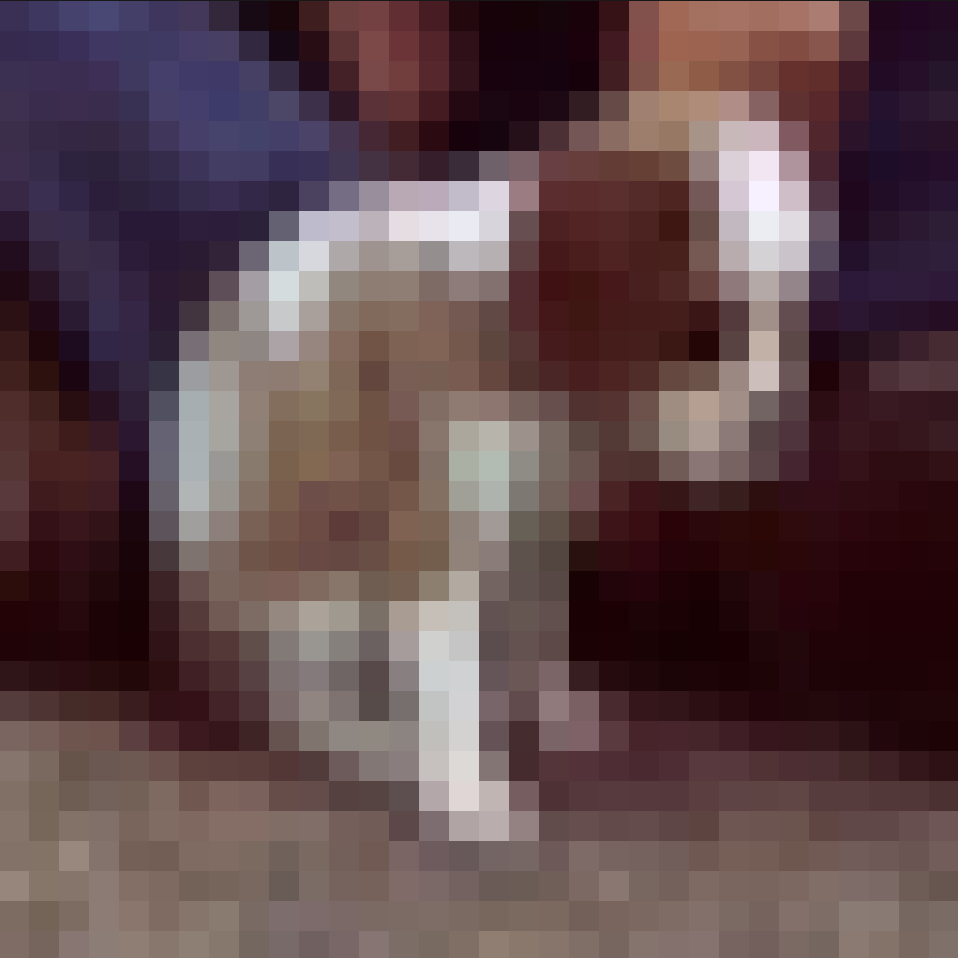}
        \subcaption{Adversarial}
    \end{subfigure}
    ~
    \begin{subfigure}{0.3\textwidth}
        \includegraphics[width=\linewidth]{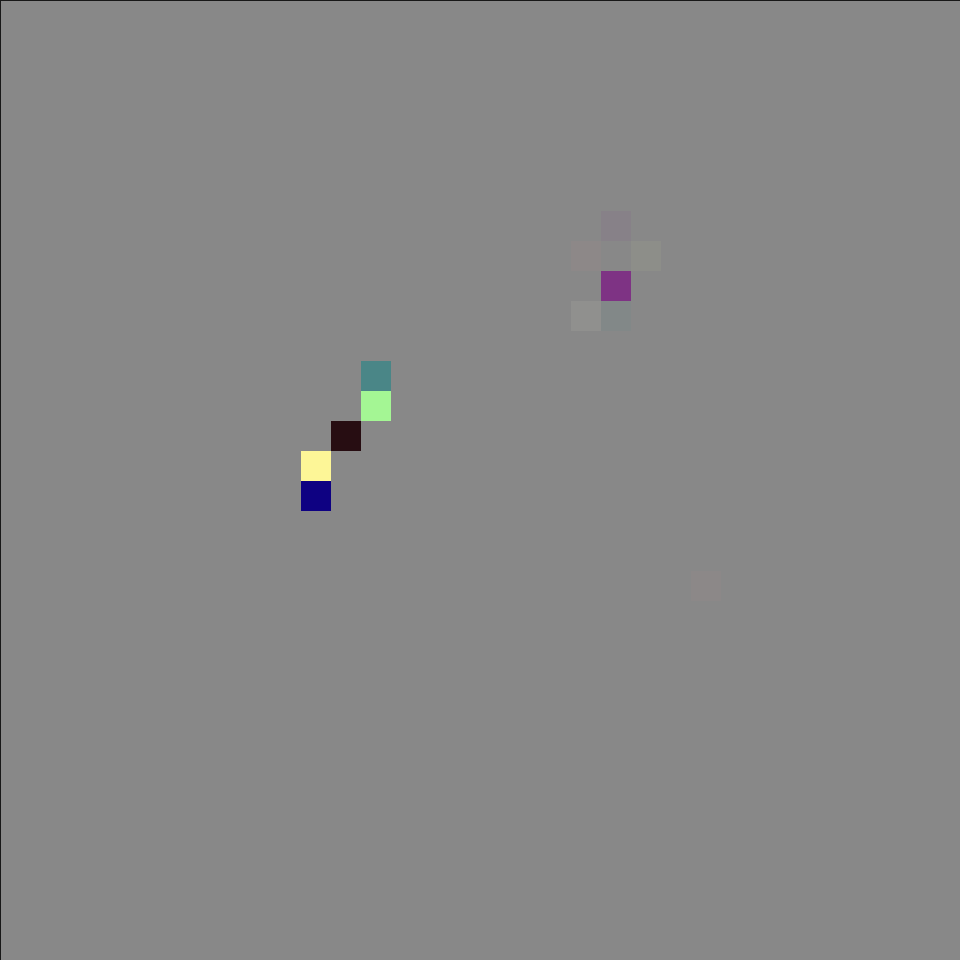}
        \subcaption{Perturbation}
    \end{subfigure}
  }
  \caption{PFW attack (15 steps), $\ell_1$-ball, $\varepsilon=1/255$, VGG-19, CIFAR-10. Left: original (“dog”); middle: adversarial (“cat”); right: perturbation.}
  \label{fig:cifar10_pfw_l_1}
\end{figure}

A notable property of FW-type methods under $\ell_1$ constraints is the production of extremely sparse perturbations. 
With $\varepsilon=64/255$ on CIFAR-10, we frequently observe successful attacks that modify a few pixels. 
This aligns with the LMO on the $\ell_1$-ball constraint, which selects the coordinate with the largest gradient magnitude.
Figures~\ref{fig:cifar10_fw_l_1}, \ref{fig:cifar10_afw_l_1}, and \ref{fig:cifar10_pfw_l_1} show examples where the adversarial perturbation affects at most a few pixels, yet flips the prediction.
This observation explains the interest in such a constraint, even though using other norms generally yields more effective attacks. 

\subsection{Logistic regression}
We start our comparison of the vanilla FW and its advanced modifications with the simple Logistic regression model and the MNIST dataset.
We focus on the $\ell_1$-ball constraint.
Table~\ref{tab:logreg_l_1} presents the test accuracy after attacks for the PGD and FW-type methods with $\ell_1$-norm ball constraint and the FGSM method that uses $\ell_{\infty}$-norm ball.
We add the FGSM method as a reference and the most popular method of first choice.
We test different magnitudes of $\varepsilon$ to highlight the dependence of test accuracy after attacks on attack strength.
We observe that the Frank-Wolfe method and its momentum version outperform PGD in both runtime and test accuracy.
At the same time, we do not observe a significant gain in the Frank-Wolfe momentum method over the vanilla version.
A possible reason for this effect is that FWm yields a sparser attack than the vanilla FW method and, therefore, its efficiency appears limited.
Last but not least, the larger $\varepsilon$ is, the smaller test accuracy after attack, which coincides with the meaning of~$\varepsilon$.   

% Under $\ell_1$ constraints, standard FGSM, PGD, FW, and FWm achieve near-zero ASR across tested $\varepsilon$ and iteration counts, whereas AFW and PFW consistently reach 11--15\% ASR with slight gains from more iterations; see .

\begin{table}[htbp]
\centering
\caption{Dependence of test accuracy for attacked data on $\varepsilon$ for the $\ell_{1}$ norm. Model: logistic regression, dataset is MNIST (10000 images), baseline test accuracy: 92.68\%. The number of iterations for each algorithm is indicated in parentheses.
We denote by * the runtimes measured in the single-image regime, in which the GPU is not fully utilized.}
% \resizebox{\textwidth}{!}{
\begin{tabular}{lcccccc}
\toprule
Algorithm & $10^{-2}$ & $5 \cdot 10^{-2}$ & $ 10^{-1}$ & $5 \cdot 10^{-1}$ & Runtime, $10^{-5}$ s. & Average nnz in $\vdelta$\\
\midrule
FGSM     & 85.56 &
16.62 &
0.44 &
0.0 &
284.86 & 476.17 \\
\midrule
PGD (1)  & 92.69 &
92.68 & 
92.68 & 
92.64 &
366.7 & 543.92\\
FW (1) &  92.63 &
92.24 &
91.88 & 
87.22 &
1.25 & 1\\
FWm (1)  & 92.63 &
92.24 &
91.88 &
87.22 &
1.16 & 1 \\
AFW (1)  & 92.65 &
92.47 &
92.24 &
90.23 &
1317.96* & 1 \\
\midrule
PGD (3)  & 92.69 & 
92.68 & 
92.66 & 
92.63 & 
685.94 & 568.74\\
FW (3) & 92.63 &
92.24 &
91.88 &
87.21 &
3.02 & 1.0127\\
FWm (3)  & 92.63 &
92.24 &
91.88 &
87.21 & 
2.9 & 1.004\\
% AFW (3) & 92.63 & 92.34 & 92.03 & 88.62 & 4439.48* & \\
% \hline
\bottomrule
\end{tabular}
% }
\label{tab:logreg_l_1}
\end{table}

\subsection{Convolutional networks}
As an example of a CNN, we consider the pre-trained ResNet-56 model and the CIFAR-10 dataset.
Table~\ref{tab:resnet_l_1} shows that FW methods generate more powerful attacks than PGD for the smaller runtime.
At the same time, we do not observe a significant gain from FWm and AFW compared to the vanilla FW method.
These methods behave similarly to the previous experiments with logistic regression and yield sparser solutions than the classical FW method.   
In addition, we note that AFW is not batch-friendly; therefore, it is difficult to fully utilize the GPU to achieve comparable runtime.

% with $\ell_1$ constraints, AFW and PFW are the only methods with non-zero ASR; AFW outperforms PFW and scales from $\sim$42\% to $>$52\% ASR as iterations increase (weak dependence on $\varepsilon$); see Table~\ref{tab:resnet_l_1}.

\begin{table}[htbp]
\centering
\caption{Dependence of test accuracy after attacks on $\varepsilon$ for the $\ell_{1}$ norm. Model: ResNet56, baseline test accuracy: 94.37\%, dataset is CIFAR10 (10000 images in the test subset). 
The number of iterations for each algorithm is in parentheses.
We denote by * the runtimes measured in the single-image regime, in which the GPU is not fully utilized.}
% \resizebox{\textwidth}{!}{
\begin{tabular}{lcccccc}
\toprule
Algorithm & $8/255$ & $16/255$ & $32/255$ & $64/255$  & Runime, $\cdot 10^{-3}$ s. & Average nnz in $\vdelta$ \\  
\midrule
FGSM & 50.72 &
41.54 &
32.6 &
20.9 &
154.03 &
3071.9 \\
\midrule
PGD (1) & 94.25 &
94.25 &
94.23 &
94.26 &
162.45 &
2936.83 \\
FW (1) & 94.0 &
93.64 &
93.23 &
92.88 &
1.27 &
1.0 \\
FWm (1) & 94.0 &
93.64 &
93.23 &
92.88 &
1.28 &
1.0 \\
AFW (1) & 94.12 &
94.0 &
93.65 &
93.23 &
250.56* &
1.0 \\
\midrule
PGD (5) & 94.25 &
94.23 &
94.23 &
94.2 &
664.66 &
2946.77 \\
FW (5) & 94.0 &
93.62 &
93.12 &
92.25 &
6.3 &
2.37 \\
FWm (5) & 94.0 &
93.61 &
93.13 &
92.23 & 
6.4 &
1.71 \\
AFW (5) & 94.01 & 93.65 & 93.16 & 92.27 & 1272.29* & 2.23\\
\midrule
PGD (10) & 94.24 &
94.23 &
94.19 &
94.19 &
1304.71 &
2956.77 \\
FW (10) & 94.0 &
93.62 &
93.09 &
92.08 &
12.79 &
2.78 \\
FWm (10) & 94.0 &
93.62 &
93.08 &
92.13 &
12.80 &
2.01 \\
AFW (10) & 94 & 93.63 & 93.06 & 92.1 & 3894.92* & 2.79 \\
% FGSM & 0.0 & 0.0 & 0.0 & 0.0 & 0.0 & 0.0 & 0.0 & 1.8 & 5.4 & 473.23 \\
% PGD (15) & 0.0 & 0.0 & 0.0 & 0.0 & 0.0 & 0.0 & 0.0 & 0.0 & 1.8 & 4424.43 \\
% FW (15) & 0.0 & 0.0 & 0.0 & 0.0 & 0.0 & 0.0 & 0.0 & 0.0 & 2.1 & 5042.24 \\
% FWm (15) & 0.0 & 0.0 & 0.0 & 0.0 & 0.0 & 0.0 & 0.0 & 0.0 & 3.1 & 5609.35 \\
% AFW (15) & 0.0 & 0.0 & 0.0 & 0.0 & 0.0 & 0.0 & 0.0 & 0.0 & 3.1 & 5801.60 \\
% PFW (15) & 0.0 & 0.0 & 0.0 & 0.0 & 0.0 & 0.0 & 0.0 & 0.0 & 1.0 & 5795.57 \\
\bottomrule
\end{tabular}
% }
\label{tab:resnet_l_1}
\end{table}

\subsection{Vision transformers}

The final model considered in our study is the Vision Transformer.
It achieves the highest test accuracy on CIFAR-10 among previous models.
Table~\ref{tab:vit_l_1} presents the test accuracy after attacks generated by the considered optimizers. 
We still observe that the vanilla FW dominates both PGD and FWm methods and provides a promising trade-off between the sparsity and attack efficiency.
In particular, after 10 iterations, the average number of nonzero elements in the perturbation is only 2.78, whereas the test accuracy is the lowest among the methods.
This observation can be used for further improvement of the FW-type methods for solving the problem of adversarial attack generation.
Moreover, the runtime of FW-type methods is moderate, and they can process multiple images simultaneously. 

% For ViT under $\ell_1$ constraints, AFW and PFW again dominate. 
% AFW improves steadily from $\sim$32\% (1 iter) to $>$50\% (15 iters), while PFW stays near $\sim$32\% across settings; see Table~\ref{tab:vit_l_1}. 
% The per-image computation time increases with the number of iterations.

\begin{table}[htbp]
\centering
\caption{
Dependence of test accuracy after attacks on $\varepsilon$ for the $\ell_{1}$ norm. Model: ViT, baseline test accuracy: 97.28\%, dataset is CIFAR10 (10000 images in the test subset). 
The number of iterations for each algorithm is in parentheses.}
% \resizebox{\textwidth}{!}{
\begin{tabular}{lcccccc}
\toprule
Algorithm & $8/255$ & $16/255$ & $32/255$ & $64/255$  & Runime, s. & Average nnz in $\vdelta$ \\  
\midrule
FGSM & 22.58 &
15.98 &
13.24 &
 12.67 & 
 0.51 & 150528
\\
\midrule
PGD (1) & 97.28 &
 97.28 &
 97.28 &
 97.28 & 0.52
 & 35752.48
 \\
FW (1) & 97.27 & 97.26
 & 97.25
 & 97.23
 & 0.38
 & 1
 \\
FWm (1) & 97.27 &
97.26 &
97.25 &
97.23 &
0.38 &
1 \\
% AFW (1) & 94.12 &
% 94.0 &
% 93.65 &
% 93.23 &
% 250.56 &
% 1.0 \\
\midrule
PGD (5) & 97.28 &
 97.28 &
 97.28 &
 97.28 &
 2.09
 & 36216.57
 \\
FW (5) & 97.27 &
97.26 &
97.23 &
97.21 &
1.94 &
1.91 \\
FWm (5) & 97.27 &
97.26 &
97.25 &
97.23 & 
1.94 &
1.45 \\
% AFW (5) & 94.01 & 93.65 & 93.16 & 92.27 & 1272.29 & 2.23\\
\midrule
PGD (10) &  97.28 &
 97.28 &
 97.28 &
 97.28 &
 4.04 & 36940.21 \\
FW (10) & 94.0 &
93.62 &
93.09 &
92.08 &
4.79 &
2.78 \\
FWm (10) & 97.27 &
97.26 &
97.25 &
97.22 &
3.89 &
1.63 \\
% AFW (10) & 94 & 93.63 & 93.06 & 92.1 & 3894.92 & 2.79 \\
\bottomrule
\end{tabular}
% }
\label{tab:vit_l_1}
\end{table}

\section{Conclusion}
\label{sec:conclusion}

This study compares projection-based and projection-free optimization methods for generating white-box adversarial attacks.
We consider three model types and two datasets in our experimental evaluation.
We compare PGD, Frank-Wolfe, Frank-Wolfe with momentum, and away steps Frank-Wolfe methods.
In addition to test accuracy after attacks, we measure perturbation sparsity to identify the underlying reasons for the observed attack performance.
The vanilla FW method demonstrates attacks that reduce the test accuracy more significantly than competitors for all models and datasets.
One possible reason is that advanced modifications produce extremely sparse perturbations, thereby reducing the attack's effectiveness.
Another potential improvement is to introduce adaptive step-size selection that does not incur prohibitive cost overhead while increasing the success rate of the resulting attack.

% Advanced Frank--Wolfe methods generate effective white-box adversarial attacks with low computational cost across diverse models (logistic regression, ResNet-56, VGG-19, ViT) and datasets (MNIST, CIFAR-10). 
% Importantly, under $\ell_1$ constraints, they can yield sparse—even single-pixel—perturbations, and outperform projection-based competitors that struggle with $\ell_1$ projections. 
% Future work includes black-box and federated settings, as well as convergence analyses in those regimes.

\section*{Acknowledgement}
The authors acknowledge support from the Russian Science Foundation, grant No.~25-41-00091.
\bibliographystyle{unsrt}
\bibliography{main}

\end{document}